# Performance Comparison and Implementation of Bayesian Variants for Network Intrusion Detection

1st Tosin Ige
Department of Computer Science
*University of Texas at El Paso*
Texas, USA

2nd Christopher Kiekintveld
Department of Computer Science
*University of Texas at El Paso*
Texas, USA

*Abstract*—Bayesian classifiers perform well when each of the features is completely independent of the other which is not always valid in real world application. The aim of this study is to implement and compare the performances of each variant of Bayesian classifier (Multinomial, Bernoulli, and Gaussian) on anomaly detection in network intrusion, and to investigate whether there is any association between each variant's assumption and their performance. Our investigation showed that each variant of Bayesian algorithm blindly follows its assumption regardless of feature property, and that the assumption is the single most important factor that influences their accuracy. Experimental results show that Bernoulli has accuracy of 69.9% test (71% train), Multinomial has accuracy of 31.2% test (31.2% train), while Gaussian has accuracy of 81.69% test (82.84% train). Going deeper, we investigated and found that each Naïve Bayes variants performances and accuracy is largely due to each classifier assumption, Gaussian classifier performed best on anomaly detection due to its assumption that features follow normal distributions which are continuous, while multinomial classifier have a dismal performance as it simply assumes discreet and multinomial distribution.

## I. INTRODUCTION

Security is indispensable and very crucial in modern information technology framework [2], [4], [5], [16], [18] and so, as we had gotten to grapple with the fact that there is no perfect system, no matter how sophisticated or state of the art a system could be, it can be attacked and compromised. With hackers constantly coming up with ever changing innovative and highly sophisticated ways to compromise system, focus had shifted to making state of the art system extremely complicated and tedious to be compromised since there cannot be a perfect system. Before any system could be compromised, there must be an intrusion for any damage to occur [1], [9], [12], [15]. It is one thing for a system to be intruded, it is another thing for the intrusion to be immediately detected and dealt with before any compromise is made. An intrusion that lasted for about fifteen (15) milliseconds before being dealt with by a combination of machine learning (to accurately detect the actual intrusion) and game theory (changing parameters and configurations to prevent further attack) approach gives an insight of a system perfection.

Naïve Bayesian algorithms are some of the most important classifiers used for prediction. Bayesian classifiers are based on probability and with general assumption that all features are independent of each other which doesn't usually hold in real world, these assumptions account for why Naïve Bayes algorithm performed poorly on certain classification, the assumption is in addition to individual assumption as each variant of Naïve Bayes classifiers of.

  i. Multinomial Naïve Bayes
 ii. Bernoulli Naïve Bayes
iii. Gaussian Naïve Bayes

Has different assumptions which impacts their efficiency and accuracy on certain tasks. Virtually all existing comparison and evaluation of Bayesian classifier with other algorithm are without acknowledgement of the fact that each variants works based on different assumption which affects their efficiency and accuracy depending on the type of classification. Since, each Bayesian variant performed differently, it sounds interesting to understand how each algorithm performed on intrusion dataset, and to understand the reason why some intrusions are not being detected by model until system compromise when the wrong Bayesian variants was being adopted.

The main contribution of our research is stated below.

- We showed that Gaussian Naïve Bayes algorithm performed best among all the three variants of Bayesian algorithm on anomalous detection in network intrusion in terms of efficiency and accuracy on KDD dataset, followed by Bernoulli with 69.9% test accuracy while Multinomial have abysmal performance with 31.2% accuracy.

- Our investigation also shows that each Bayesian algorithm works based on its assumption regardless of data. Gaussian Naïve Bayes performed better on anomaly detection in network intrusion because of its assumption that features follow normal distributions which are continuous. So, we are sure that the algorithm factored in all target categories.

## II. RELATED WORK

### a. MULTINOMIAL NAÏVE BAYES

In multinomial Naïve Bayes, features are assumed to be from a multinomial distribution [3], [6], [17] which gives the probability of observing counts from a few categories [13], [14], this makes it very effective for prediction when the feature(s) is discrete and not continuous.

The likelihood of **S** given **C**$_k$ is the product of terms' probabilities **p**$_{ki}$ in the statistical degree of $w_{ki}$, thereby rejecting null hypothesis:

The probability of document **d** being in class **c** being computed as;

$$P(c|d) \alpha\ P(c)\ \pi_{1<k<n}\ P(t_k|c) \quad (1)$$

Where **P(t$_k$|c)** is the conditional probability of $t_k$. **P(t$_k$|c)** is interpreted as a measure of evidence contributed by $t_k$ to the fact that **C** is the correct class. **P(c)** is the prior probability of a document occurring in class **C**. If a document's terms do not provide clear evidence for one class versus another, we choose the one that has a higher prior probability. ($t_1, t_2, t_3 \ldots, t_{nd}$). The best class in Bayesian classification is the most likely or maximum posteriori (MAP) class.

$$C_{map} = \text{argmax}_{c \in c} p(c|d) = \text{argmax}_{c \in c} p(c) \pi_{1<k<n}\ P(t_k|c) \quad (2)$$

**P(c)** and **P(t$_k$|c)** are being estimated from the training set as we will see in a moment. Since conditional probabilities are being multiplied together, one for each position 1<k<nd. This can result in an underflow of floating points. And so, it is better for us to do the computation by adding logarithms of probabilities instead of multiplying the actual probabilities. Hence, the class with the higher logarithmic probability will still be the most probable. The class with the highest log probability score is still the most probable which is;

**log(xy) = log(x) + log (y).** Hence,

$$C_{map} = \text{argmax}_{c \in c}[\log p(c) + \Sigma_{1<k<nd} \log P(t_k|c)] \quad (3)$$

, is the actual maximization that is being done in the implementation of Naïve Bayes which is the maximization of the log to get the correct class of the input. Multinomial Naïve Bayes is good for training a model when we have discrete variable and when the distribution is multinomial in nature. The assumption that the distribution is multinomial coupled with additional assumption of independence among the features makes multinomial Naïve Bayes a drawback when the two assumptions are not valid in test or train data.

### c. GAUSSIAN NAÏVE BAYES

As a typical Bayesian classifier which assumed the value of each feature to be completely independent of the other, it assumed that each continuous value in a continuous data is distributed according to Gaussian distribution [7],[8]. Hence, the probability of individual features is assumed to be.

$$P(X_1 | Y) = \frac{1}{\sqrt{2\pi\sigma_{2y}}} \exp(-((x_i - u_y)_2)/ 2\sigma_{2y}) \quad (4)$$

For which all parameters are independent of each other. One of the simplest ways to approach this is to assume that the data has Gaussian distribution without any co-variance. All we have to do is to find the mean and standard deviation between each label to form our model. With perfect knowledge that our variable is normally and continuously distributed from $-\infty < X < +\infty$. The total area under the model curve is 1.

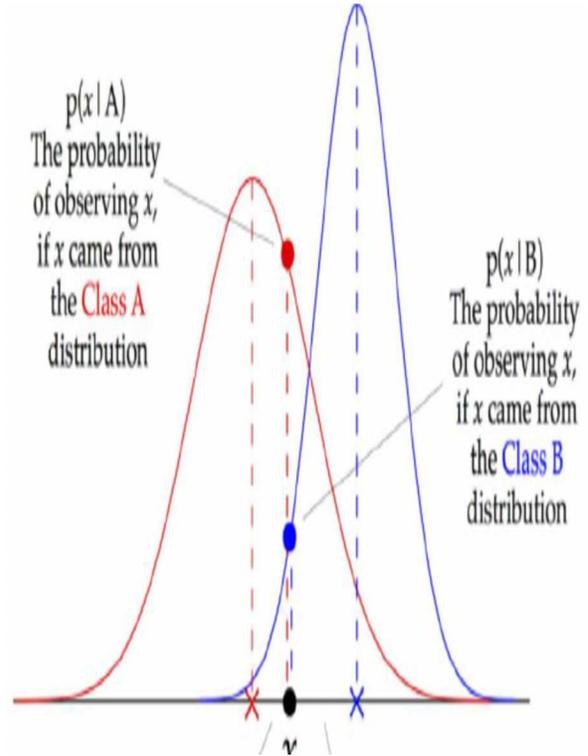

Figure 1 Probability under Gaussian Distribution

### b. BERNOULLI NAÏVE BAYES

Bernoulli Naïve Bayes [5], [10], [11] assume data is discrete and the distribution is of Bernoulli mode. Its main feature is the acceptance of binary values such as yes or no, true of false, 0 or 1, success or failure as input. Its assumption of discrete and Bernoulli distribution as

$$P(x) = P[X=x] = (q=1-p\ x=0\ p\ x=0) \quad (5)$$

Where x can be either 0 or 1 but nothing else makes it suitable for binary classification as its classification rule is according to

$$P(x_i | y) = p(1 | y)x_i + (1 - p(i | y))(1 - x_i) \quad (6)$$

Its computation is based on binary occurrence information (Figure 2), and so neglects the number of occurrences or frequency, this makes Bernoulli unsuitable for certain tasks such as document classification.

```
TRAINBERNOULLINB(C, D)
1  V ← EXTRACTVOCABULARY(D)
2  N ← COUNTDOCS(D)
3  for each c ∈ C
4  do N_c ← COUNTDOCSINCLASS(D, c)
5      prior[c] ← N_c/N
6      for each t ∈ V
7      do N_ct ← COUNTDOCSINCLASSCONTAININGTERM(D, c, t)
8          condprob[t][c] ← (N_ct + 1)/(N_c + 2)
9  return V, prior, condprob

APPLYBERNOULLINB(C, V, prior, condprob, d)
1  V_d ← EXTRACTTERMSFROMDOC(V, d)
2  for each c ∈ C
3  do score[c] ← log prior[c]
4      for each t ∈ V
5      do if t ∈ V_d
6          then score[c] += log condprob[t][c]
7          else score[c] += log(1 − condprob[t][c])
8  return arg max_{c∈C} score[c]
```

Figure 2: Bernoulli Algorithm Stepwise

Single occurrence of the word geography in physics textbook can make it to be classified as geography as Naïve Bayes algorithm doesn't factor the number of occurrences unlike multinomial.

### III. RESEARCH METHODOLOGY

#### a. FEATURE SELECTION

Feature selection, which is also being referred to as attribute selection is a process of extracting the most relevant features from the dataset for the purpose of using classifier to train a model in order to ensure overall better performance of the model. Since the presence of large number of irrelevant features in dataset increases both the training time and the risk of overfitting, having an effective feature selection method is a necessity. We use Chi-square test for categorical features in KDD dataset. The calculation in Chi-square test between each feature and the target greatly helps to determine if there is any association between two categorical variables in the dataset and whether such association will influence the prediction. This ultimately helps selecting the desired number of features with best Chi-square test scores. Chi-square test is a technique to determine the relationship between the categorical variables. The chi-square value is calculated between each feature and the target variable, and the desired number of features with the best chi-square value was selected.

There are three main types of Chi-square tests, tests of goodness of fit, the test of independence, and the test for homogeneity. All three tests rely on the same formula to compute a test statistic. Since unsupervised feature selection techniques tend to ignore target variable like in the case of methods removing redundant variables using correlation. We chose supervised feature selection techniques which use target variables like methods that can remove irrelevant variables from dataset.

```
In [16]: #Correlation with output variable
         cor_target = abs(cor["Label"])
         #Selecting highly correlated features
         relevant_features = cor_target[cor_target>0.5]
         relevant_features

Out[16]: Bwd_Packet_Length_Max      0.648457
         Bwd_Packet_Length_Mean     0.656620
         Bwd_Packet_Length_Std      0.654563
         Flow_IAT_Std               0.560958
         Flow_IAT_Max               0.611038
         Fwd_IAT_Std                0.615898
         Fwd_IAT_Max                0.611497
         Max_Packet_Length          0.626795
         Packet_Length_Mean         0.602986
         Packet_Length_Std          0.641535
         Packet_Length_Variance     0.603295
         Average_Packet_Size        0.596895
         Avg_Bwd_Segment_Size       0.656620
         Idle_Mean                  0.614349
         Idle_Max                   0.614669
         Idle_Min                   0.609732
         Label                      1.000000
         Name: Label, dtype: float64
```

Fig 3 CHI Square Feature Selection Test result

#### b. EXPERIMENTAL SETUP

We did separate implementation for each Bayesian variants (Gaussian, Multinomial, and Bernoulli) using KDD dataset obtained from Kaggle. It has (692703, 79) shape, the dataset could not be used directly without preprocessing due to the presence of several nil values, missing values, and negative value in it. Our task of preprocessing the dataset involves the following step.

1. Visualizing the list of categorical variables in the dataset

2. Viewing frequency counts and distribution of each categorical variable

3. Checking for missing value in the categorical variable

4. Checking for missing value and visualization of the numerical variable

Altogether, we have thousands of NAN values, NULL values, and negative values. So, we adopted statistical method of approach that is based on the use of standard deviation and mean to analyze each column. There are 79 columns in the dataset and so, it is wise to treat each column as separate entity when addressing missing values, hence for individual column, a combination of standard deviation and mean for the column was used to fill the missing and null values. There are six categories in the label dataset which needs encoding, and so one-hot encoding was used to encode the categorical variable. One of the most important tasks in the data preprocessing was the feature selection (Figure 3), since our dataset has 79 features, only relevant features are needed to get optimal result.

A combination of CHI square feature selection test method and confusion matrix was used to select only the relevant features in the dataset to train our model while the irrelevant features were ignored. Our implementation gave separate results and observation for each variant of Naïve Bayes. Multinomial gave abysmal performance on the KDD dataset and was the worst performance among the three (Figure 4), (Figure 5), (Figure 6).

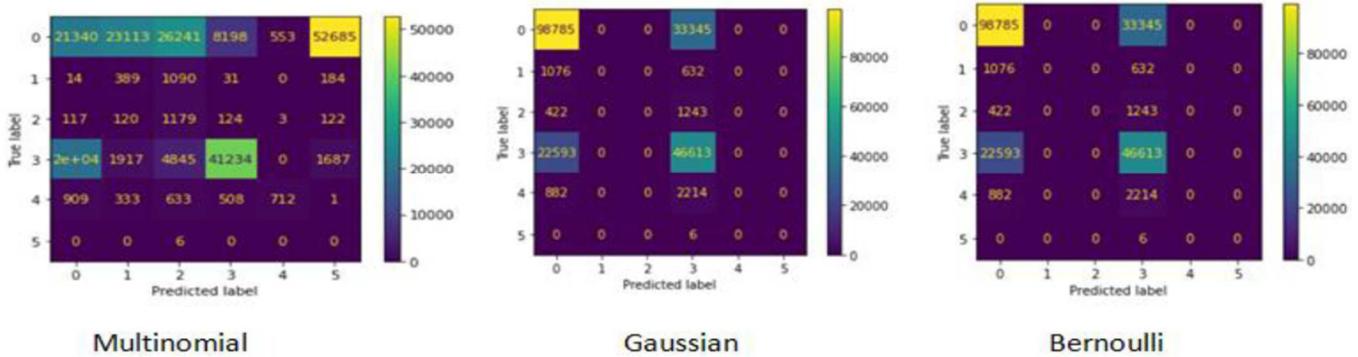

Figure 4. Confusion matrix of each Bayes on Security Dataset

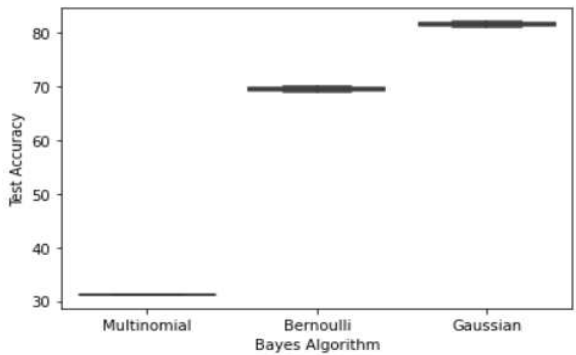

Figure 5. Box plot performance comparison

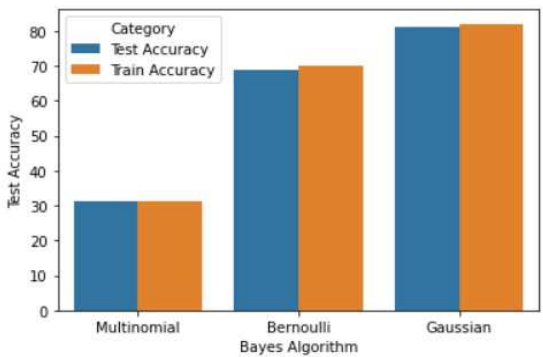

Figure 6. Variation in Train and Test Performance in Variant Bayesian Algorithm

It was important to visualize test and train accuracy, so as to ensure the classifier doesn't memorize the training data as that could cause a very wide gap between train and test accuracy which is over fitting, as seen in (figure 6) there is closeness between train and validation accuracy to validate the implementation. Experimental result shows that, for each variant of Bayesian algorithm, Bernoulli have accuracy of 69.9% test (71% train), Multinomial has accuracy of 31.2% test (31.2% train), and Gaussian has accuracy of 81.69% test (82.84% train) (Figure 5), (Figure 6) on security dataset. Comparing each variant accuracy and performance will be incomplete without diving on what might be responsible for differences in the performances of each Bayesian variant. We went back to thoroughly re-analyzed each of the relevant features and categorical label for any observation and then compare whatever our observation is to each variant assumption since each of the three variants of Bayesian algorithm has different assumption. Our label has six categories of attack ['BENIGN', 'DoS slowloris', 'DoS Slowhttptest', 'DoS Hulk', 'DoS GoldenEye', 'Heartbleed'] which are all encoded between 0 and 5. There was no relevant observation in the dataset except the label categories which ranges from 0 to 5 from which predicted output are continuously selected from. This is according to normal distribution and explains why Gaussian variants of Bayesian classifier work best for this implementation. The fact that the label distribution is not discrete but continuous clearly indicates why multinomial naïve Bayes have abysmal performance. Bernoulli have mixed performance in between because it takes the top two from the label category and based its prediction label as it assumes binary classification.

## IV. CONCLUSION

We concluded that the performance of each variant of Bayesian classifier is impacted by its assumption, and each assumption is the single most important factor that influences their performance and accuracy, as we can see in the normal and continuous distribution of our label category which causes Gaussian Bayes to work best among Bayesian variant on security dataset. If the dataset is discrete like document classification, we can expect multinomial to work best and Bernoulli to work best for binary classification such as true/false, yes/no and so on. We showed that Gaussian Naïve Bayes algorithm performed best among the entire Bayes based algorithm on intrusion detection, we also showed that each variant of Bayes algorithm blindly follows its assumption which is the single most important factor that influences their performance.